\documentclass[letterpaper, 10 pt, conference]{ieeeconf}  

\IEEEoverridecommandlockouts                              

\usepackage{booktabs} 
\usepackage{CJKutf8}
\newcommand{\ja}[1]{\begin{CJK}{UTF8}{ipxg}#1\end{CJK}}

\usepackage[hidelinks]{hyperref}

\usepackage{graphicx}   
\usepackage{overpic}    
\usepackage{amsmath}    
\usepackage{amssymb}    
\usepackage{tikz}
\usetikzlibrary{arrows.meta, calc}
\usepackage{tcolorbox}

\definecolor{wycolor}{RGB}{0,120,140}
\newcommand{\wy}[1]{#1}

\usetikzlibrary{positioning}
\title{\LARGE \bf
Emotion Recognition in Sign Language Conversation
}

\usepackage{fancyhdr}
\author{Yusong Wang$^{1}$, Keyu Mao$^{1}$, Takao Obi$^{1}$, Minghao Shao$^{2}$ and Kotaro Funakoshi$^{1}$
\thanks{$^{1}$ Yusong Wang, Keyu Mao, Takao Obi, and Kotaro Funakoshi are with Institute of Science Tokyo, 
Japan.
        {\tt\small \{wangyi, maokeyu, obi, funakoshi\}@lr.first.iir.isct.ac.jp}}
\thanks{$^{2}$ Minghao Shao is with New York University, New York, 11201, USA.
        {\tt\small shao.minghao@nyu.edu}}
}

\begin{document}

\maketitle
\thispagestyle{fancy}

\begin{abstract}
Emotion Recognition in Conversation is a core component of affective computing, while current sign language emotion datasets primarily focus on isolated sentences and lack conversational context. 
Models trained exclusively on these isolated utterances demonstrate degraded performance in real world scenarios because they cannot utilize historical dialogue flow. 
To address this structural limitation, we introduce the ERC task to sign language video analysis and propose the eJSL Dialog dataset. 
Constructed using the scripts from the STUDIES corpus, the dataset contains 1,920 video samples organized into 480 unique dialogues. 
\wy{We conduct systematic benchmarking on this dataset using models ranging from isolated visual networks to multimodal conversational architectures.
The results suggest the feasibility of extending conversational ERC frameworks to sign-language dialogue under the current benchmark setting, while also revealing limitations in existing visual representations for capturing sign-specific affective cues, motivating future work on sign-specific visual modeling and larger sign-language conversational training resources.}
\end{abstract}

\pagestyle{plain}
\section{INTRODUCTION}
Emotion recognition enables machines to perceive users' affective states and provide empathetic responses, making it central to virtual assistants and emotionally aware assistive technologies \cite{picard2000affective,poria2017review,wu2025comprehensive}.
Most systems focus on spoken languages and standard facial expressions from non-signing populations. 
Extending these technologies to sign language, which is often ignored in mainstream research, is a necessary step \cite{koller2020quantitative, al2021deep}.
In sign language communication, the emotion recognition task becomes highly complex due to the visual nature of the language itself. 
As visual languages, sign languages rely on hand signs,
facial expressions, and upper body movements to convey linguistic structure and emotional content simultaneously \cite{corina1999neuropsychological,elliott2013facial}. 
This overlap between grammatical features and emotional features introduces significant ambiguity to automatic emotion recognition systems \cite{corina1999neuropsychological, freitas2017grammatical}. 
Accurately capturing and modeling emotional dynamics in sign language remains a practical challenge in the field of both computer vision and language processing.

Existing resources for sign language emotion analysis focus primarily on isolated sentences or unidirectional expressions. 
For instance, datasets such as eJSL Solo \cite{funakoshi2025emotion} consist of sign language video clips detached from conversational context. Similarly, the EmoSign dataset \cite{chua2025emosign} concentrates on capturing emotional expressions within single video utterances. 
These datasets have advanced research in isolated sign language emotion recognition, but they omit the conversational context present in real communication.

In general, emotion recognition models trained exclusively on these isolated utterances cannot utilize historical context. 
Consequently, they degrade performance when applied to real world scenarios where emotional meaning depends on the continuous dialogue flow \cite{hazarika2018icon,majumder2019dialoguernn,ghosal2020cosmic}. 
In authentic bidirectional interactions, emotional states undergo a dynamic evolution process.
An individual's emotional shift is influenced by their own emotional history and also occurs as a direct response to the state of their interlocutor.
The absence of dialogue history limits the ability of existing models to understand complex emotional evolution. Therefore, exploring emotion recognition in sign language in multiple turn dialogue scenarios is a core step to advance this field.

To address this structural limitation and the lack of dialogue data in existing research, we propose the eJSL Dialog dataset for the Emotion Recognition in Conversation (ERC) task in sign language. 
This represents the first exploration into this specific task. 
The eJSL Dialog dataset is constructed using the dialogue scripts from the STUDIES Japanese Empathetic Dialogue Speech Corpus \cite{saito2022studies}. 
Each line of the scripts has a designated emotion category label.
The constructed dataset contains a total of 1,920 video samples divided into 480 unique dialogues centered around teacher and student interactions. 
\autoref{tab:dialog_example} presents a dialogue example from STUDIES, illustrating the dynamic emotional exchange between the student and the teacher.

\begin{table}[t]
\centering
\renewcommand{\arraystretch}{1.4}
\caption{Example of dialogue lines from STUDIES \cite{saito2022studies} by a teacher and a male student used in our eJSL Dialog dataset. 
}
\label{tab:dialog_example}
\tabcolsep=1mm
\begin{tabular}{l l p{5.8cm}}
\hline\hline
\textbf{Speaker} & \textbf{Emotion} & \textbf{Line} \\
\hline
Male & Joy & \ja{先生！この前部活の試合で勝ったんだ！} \\
student & & (Teacher! I won my club match the other day!) \\
\hline
Teacher & Joy & \ja{文武両道だね！} \\
& & (You are excelling in both academics and sports!) \\
\hline
Male & Joy & \ja{そう！それを目指してる！} \\
student & & (Yes! That is what I am aiming for!) \\
\hline
Teacher & Neutral & \ja{あなたなら出来るわ。これからもしっかり頑張るのよ！} \\
& & (You can do it. Keep working hard from now on!) \\
\hline\hline
\end{tabular}
\end{table}

To establish an objective evaluation benchmark, we applied and compared multiple baseline methods on this dataset.
These models span purely visual emotion recognition networks, a text based conversational emotion recognition model, and multimodal conversational emotion recognition architectures.
\wy{Our benchmark results suggest the feasibility of extending conversational ERC frameworks to sign-language dialogue under the current evaluation setting. 
At the same time, the results reveal limitations in existing visual representations for capturing sign-specific affective cues, indicating the need for more effective sign-specific visual modeling and larger-scale sign-language conversational training datasets to support such model development.}

The main contributions of this paper are as follows:

\begin{itemize}

\item We formally define the ERC task for sign language video analysis to establish an objective evaluation benchmark for bidirectional interaction scenarios.

\item We construct and release the eJSL Dialog dataset, providing sign language video samples with explicit multiple turn dialogue context and corresponding emotion annotations to address the structural limitations of isolated utterance datasets.

\item We conduct systematic benchmarking on this dataset using models.  
We demonstrate the limitations of isolated visual models and generic multimodal conversational emotion recognition models, confirming the explicit need for context aware visual extractors specific to sign language and indicating that expanding the scale of conversational datasets to support large scale pre-training is a necessary next step.

\end{itemize}

\section{BACKGROUND}

\subsection{Sign Language \& Emotion}
Sign language utilizes visual-manual modality and non-manual markers, such as facial expressions and upper body movements, to convey information \cite{sandler2006sign,brentari2010sign}. 
A key challenge in sign language emotion recognition is that facial expressions serve dual purposes \cite{corina1999neuropsychological, freitas2017grammatical}. 
They encode both grammatical structures, such as questions, and affective states, such as surprise. 
This overlap introduces ambiguity for automatic recognition systems because the same facial movement can indicate either a linguistic function or an emotional response \cite{silva2020recognition}.
Indeed, the research on emotion in sign language recognition is quite scarce.
For example, while a survey~\cite{koller2020quantitative} refers to over 200 relevant papers, no emotion recognition work is included in the survey.

\subsection{Visual Emotion Recognition}
Current methods address visual emotion recognition by integrating facial and hand gesture features into multimodal frameworks \cite{gu2023wife,wei2024learning,10024142}. 
However, these approaches primarily focus on classifying isolated video sequences \cite{zhang2023weakly,zhang2024mart,xue2024affective}. 
They typically rely on frame level spatial feature aggregation and short term temporal tracking. 
Consequently, they do not possess the structural capacity to model the long term emotional dynamics and conversational context dependencies present in continuous and interactive communication.

\subsection{Emotion Recognition in Conversation}
ERC involves identifying the affective states of participants across multiple turns in a dialogue \cite{poria2017review,pereira2024deep}.
In realistic interactions, emotion evolves based on personal history and interlocutor responses \cite{hazarika2018icon,majumder2019dialoguernn}.
Current ERC research primarily focuses on spoken languages using textual and acoustic modalities. 
These frameworks often include visual cues, they are optimized for non signing populations where facial expressions predominantly reflect affective states \cite{gu2023wife,wei2024learning,10024142}.
As explained, in sign language, facial expressions serve dual roles by encoding grammatical structures and emotional content simultaneously. 
\wy{This difference may create a domain mismatch for existing visual representations, since sign-language visual cues jointly encode linguistic and affective information.}
Exploring ERC in sign language is necessary to model dynamic transitions in actual communication.

\subsection{Existing Emotion Sign Language Datasets}
Existing datasets for sign language emotion recognition consist of isolated sentences or single expressions. 
For example, the eJSL Solo dataset \cite{funakoshi2025emotion} contains individual video clips of signers performing specific sentences under instructed emotion categories in Japanese Sign Language (JSL). 
Similarly, the EmoSign dataset \cite{chua2025emosign} focuses on capturing affective expressions within single American Sign Language video utterances. 
However, because these recordings are detached from any conversational flow, they lack the multiple turn dialogue history required for conversational analysis. 
The absence of sequential interaction means these datasets cannot represent how a signer adjusts their emotional expression based on the previous statements of a partner. 
This limitation prevents researchers from developing models that understand context dependent emotional changes in sign language.

To address the limitations of existing datasets and the absence of conversational context in current research, we construct the eJSL Dialog dataset. 
\autoref{tab:dataset_comparison} compares EmoSign, eJSL Solo and eJSL Dialog. eJSL Solo and Dialog are perfectly and nearly balanced respectively but both are not spontaneous (acted), while smaller EmoSign is spontaneous but not balanced.

\begin{table*}[t]
\centering
\renewcommand{\arraystretch}{1.2}
\setlength{\tabcolsep}{8pt}
\caption{Comparison of existing sign language datasets with emotion annotations.}
\label{tab:dataset_comparison}
\begin{tabular}{l c c c c c c c}
\toprule
\textbf{Dataset} & \textbf{Language} & \textbf{Interaction Type} & \textbf{Size} & \textbf{Signers} & \textbf{Emotion Labels} &
\textbf{Balanced} & \textbf{Spontaneous}\\
\hline
EmoSign & ASL & Isolated Utterances & 200 clips & 4 & 10 classes & $\times$ & $\checkmark$ \\
eJSL Solo & JSL & Isolated Utterances & 1,092 clips & 2 & 7 classes &$\checkmark$ & $\times$\\
eJSL Dialog (Ours) & JSL & Multi-Turn Dialogues & 1,920 clips & 2 & 4 classes &$\checkmark$ & $\times$\\
\bottomrule
\end{tabular}
\end{table*}

\section{METHODS}

\subsection{Data Source and Script Selection}
We derived the linguistic content for our dataset from the short dialogue subset of the STUDIES corpus \cite{saito2022studies}. 
In the original corpus, the short dialogue scripts were collected through a microtask crowdsourcing mechanism. Specifically, the data collection process involved launching 12 microtasks and recruiting 100 participants for each task. After an initial screening, this process yielded a total of 720 short dialogue texts with 4 emotion types. 
Every single utterance within these dialogues is equipped with an explicit emotion label.
The original text scripts underwent manual revision to remove typographical errors and inappropriate expressions, yielding a clean text baseline for sign language adaptation. 

From this collection, we selected 480 dialogues to construct our dataset. 
The selection was aimed to balance instances across four emotion types and binary genders.
Each dialogue consists of four consecutive utterances.
This length provides sufficient conversational history to model emotional transitions and is suitable for a simplified evaluation of empathetic dialogue systems.

\begin{figure}[t]
\includegraphics[width=0.45\textwidth]{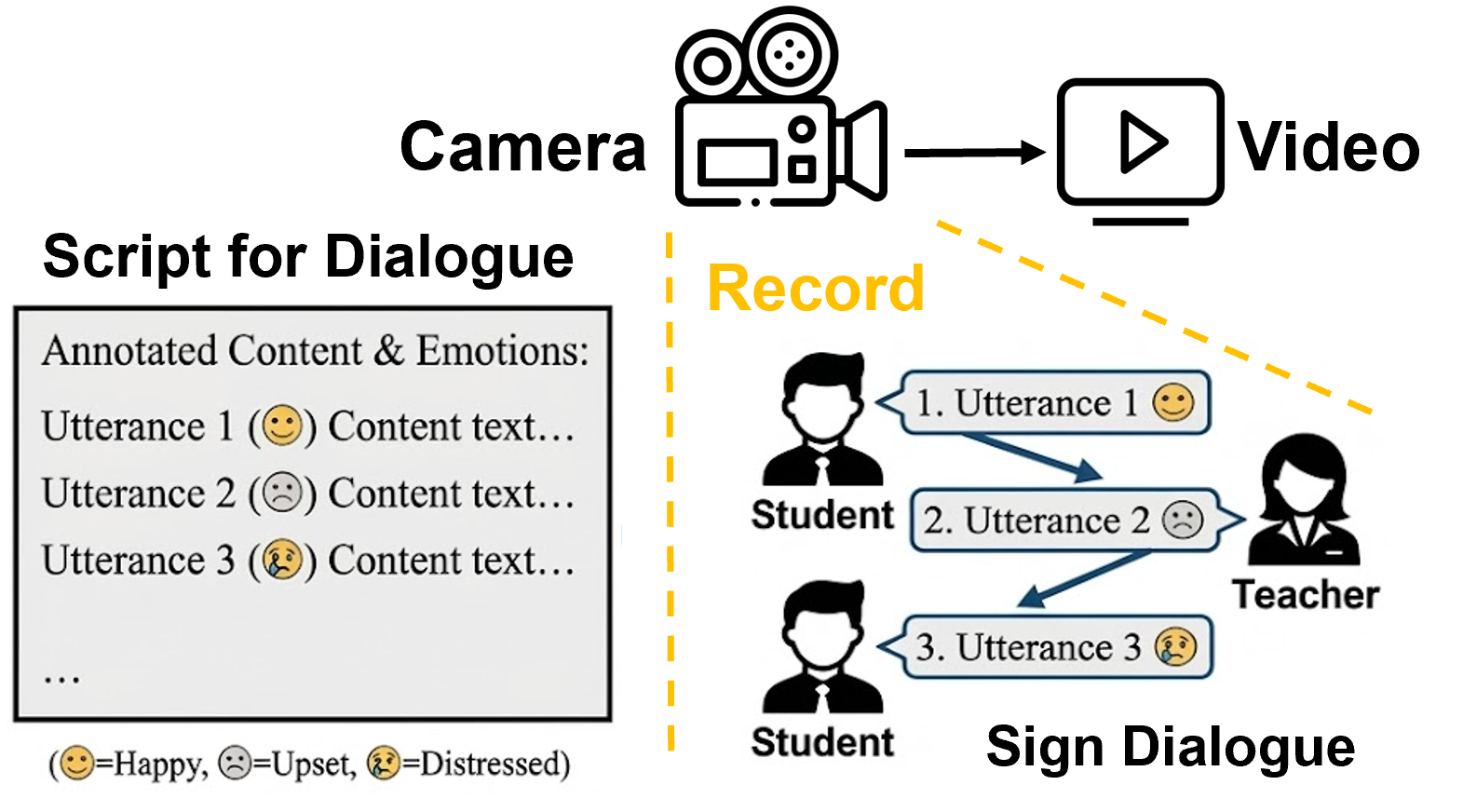}
\centering
\caption{Illustration of the sign language video recording process. The actors perform the dialogue by sign language based on the provided script and emotion labels. An RGB camera for each actor records interleaved performances to produce the final video dataset.}
\label{fig:recording_pipeline}
\end{figure}

\subsection{Sign Language Video Recording}
\autoref{fig:recording_pipeline} illustrates the recording pipeline.
We recorded the sign language videos using the selected scripts. 
The dialogues are set in a tutoring school, involving a female teacher interacting with a male or female student. 
Following the original corpus, the male student was acted by a male signer and the female student was acted by a female signer.
On the other hand, the teacher role was acted by both of them because there were only one male and one female signers.
For each dialogue line, the actors were provided with the text and the corresponding utterance level emotion labels on computer screens simultaneously, and only the corresponding actor made a recording for the line interchangeably.
Based on these instructions, the actors expressed the semantic meaning of the text and the assigned emotional state in JSL. 

The participated actors are native JSL signers who work as vocational deaf actors. 
They read and write fluently in Japanese, so all instructions and utterances were textually presented in Japanese.
The recording was conducted in 2025. 
We obtained the explicit consent of the signers using standard consent forms of the institution, and the signers were paid for their participation.
Ethical reviewing was exempted based on the prescreening of the institution.

The videos were recorded in a controlled indoor environment featuring a pure white background to eliminate visual distractions. 
We used a single RGB camera positioned at a fixed height for the recording process of each signer. 
The final videos are processed at a resolution of $1440 \times 1080$ at 30 fps. 
Each resulting clip is a complete JSL utterance conveying a single intended emotion.


\section{eJSL Dialog STRUCTURE}

\subsection{Dataset Configuration}
The eJSL Dialog dataset comprises 1,920 video clips into 480 unique dialogues. 
Spanning eight distinct scenes with exactly 60 dialogues per scene, the dataset is strictly structured so that each dialogue consists of four consecutive utterances. 
The total duration of the recorded video data is approximately 4.65 hours. 
At the utterance level, the clips have an average duration of 8.73 seconds, ranging from 2.94 to 43.98 seconds. 
The accompanying Japanese text transcripts contain a total of 134,416 characters, with an average of 70.0 characters per utterance. 
All video files are stored in mp4 format. 
We adopted a standardized file naming convention that explicitly encodes the scene identifier (SDnn), dialogue number (01--60), utterance sequence (01--04), and the corresponding emotion label for each line. 
Specific examples of this naming structure include SD09-38-03A.mp4 and SD10-26-03S.mp4.
Emotion labels consist of A: Angry, J: Joy, N: Neutral, and S: Sad.

eJSL Dialog is exclusively designed as a benchmark dataset.
It is intended strictly for the evaluation and testing of emotion recognition models and is not expected to be used for training purposes.
This decision ensures that the dataset serves as a gold standard for assessing the generalization capabilities of models trained on other resources.

\subsection{Emotion Categories and Annotation}
\begin{figure}[t]
    \centering
    \includegraphics[width=0.99\linewidth]{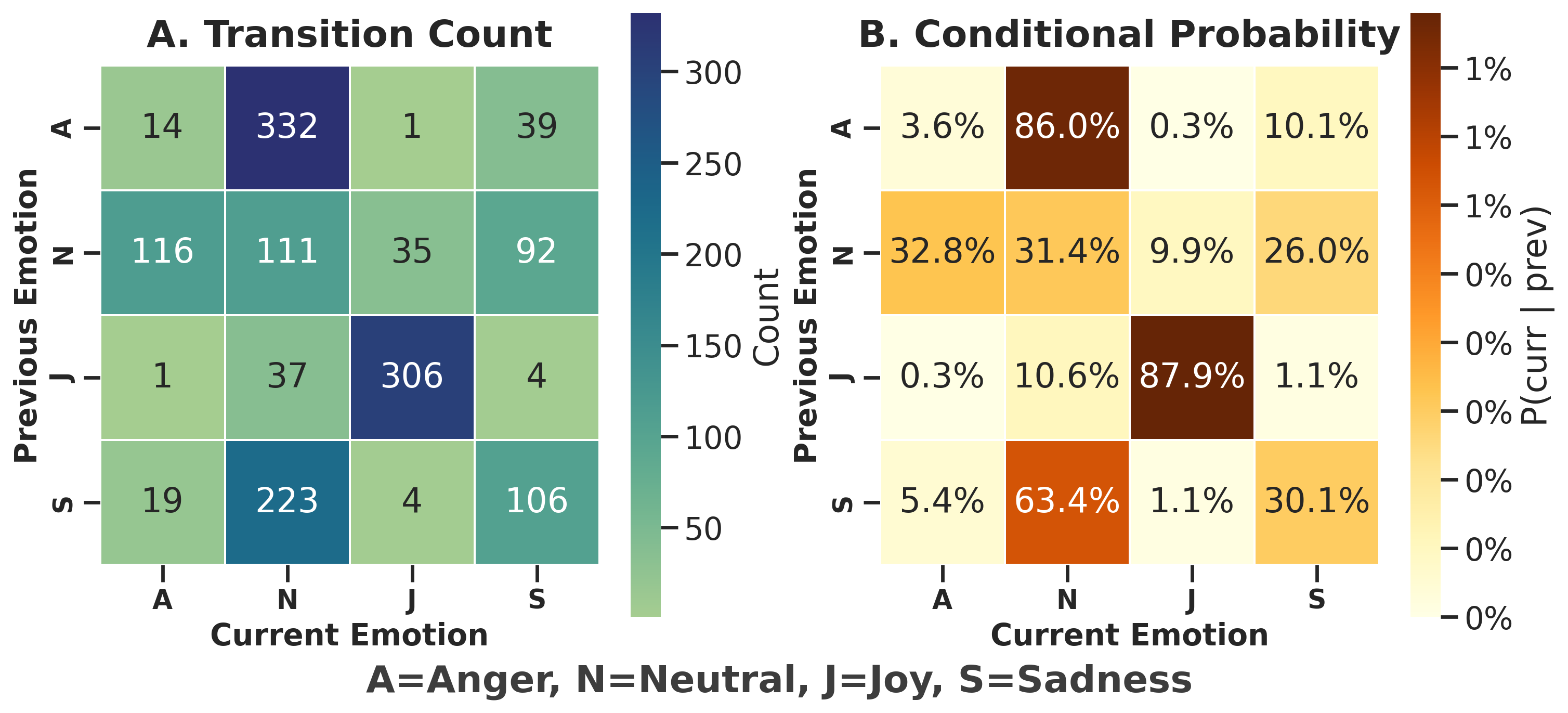}
    \caption{Emotion transition structure on eJSL Dialogue.
Panel (A) reports absolute transition counts between the previous and current utterance labels.
Panel (B) shows the corresponding row-normalized transition probabilities.}
\label{fig:ejsl_transition}
\end{figure}

Each video sample in the dataset is explicitly annotated with one of four emotion categories: Neutral, Joy, Sad, and Angry. 
\autoref{tab:emotion_dist} details the quantitative distribution of these labels across the dataset. 
The higher proportion of Neutral labels aligns with the designated conversational setting, where the teacher predominantly responds in an objective state.
The student utterances present a relatively balanced distribution across all four emotion categories.
Moreover, \autoref{fig:ejsl_transition} summarizes the  emotional dynamics of eJSL Dialog dataset. The count matrix (Panel A) reflects the empirical transition frequencies between emotion pairs, while the probability matrix (Panel B) normalizes each source row to highlight the conditional transition patterns independent of source class frequency.
This structural characteristic reflects empathetic dialogue dynamics.
We further validated the scripted labels by asking three non-signer observers to classify 40 randomly sampled clips from video only, with 10 clips per emotion class.
They achieved an average macro-F1 of 69.41\% against the scripted labels, with a Fleiss' $\kappa$ of 0.580, indicating moderate inter-observer agreement.

\begin{table}[t]
\centering
\caption{Distribution of utterance level emotion labels in the eJSL Dialog dataset.}
\renewcommand{\arraystretch}{1.1} 
\setlength{\tabcolsep}{12pt}     
\begin{tabular}{lcc}
\toprule
\textbf{Emotion Label} & \textbf{Number of Clips} & \textbf{Percentage} \\
\midrule
Neutral & 704 & 36.7\% \\
Joy & 465 & 24.2\% \\
Angry & 390 & 20.3\% \\
Sad & 361 & 18.8\% \\
\midrule
Total & 1,920 & 100\% \\
\bottomrule
\end{tabular}
\label{tab:emotion_dist}
\end{table}

\section{PROBLEM FORMULATION}
\label{sec:problem_formulation}

We formalize the emotion recognition in sign conversation task. 
Given a dialogue sequence $D = \{u_1, u_2, \dots, u_N\}$ consisting of $N$ consecutive utterances, each utterance $u_i$ contains visual modality features $V_i$ and textual modality features $T_i$. 
Here $V_i$ conveys the original sign language information and $T_i$ is an utterance-level spoken language translation or a word-for-word translation (i.e., gloss) of $V_i$.
The objective of the model is to learn a mapping function $f$ to predict the emotion label $y_t \in E$ of the target utterance $u_t$, where $E$ represents the predefined set of emotion categories. The complete prediction process relies on the current multimodal input $u_t$ and the historical context set $C = \{u_1, \dots, u_{t-1}\}$. The task is formally expressed as:
$$y_t = f(V_t, T_t, C)$$
Note that this formulation defines the comprehensive multimodal conversational task. 
Specific evaluation models may operate on a constrained subset of these inputs. 
For example, traditional visual recognition models omit text and context variables, reducing the mapping to $y_t = f(V_t)$. 
Text based conversational models omit visual features, operating as $y_t = f(T_t, C)$.
Additionally, isolated multimodal models ignore historical context information entirely, relying solely on the current utterance to predict the emotion as $y_t = f(V_t, T_t)$.

\section{BASELINE EVALUATION}

\subsection{Baselines}
To establish an objective benchmark for the eJSL Dialog dataset and analyze the roles of different modalities and historical context, we evaluate five emotion recognition models:
\begin{itemize}

\item EmoAffectNet is a purely visual emotion recognition model that processes frame level facial features \cite{RYUMINA2022}.

\item EANwH is an extension of EmoAffectNet that concatenates facial and hand skeletal features at the frame level and utilizes an LSTM network to capture temporal emotional dynamics across the video sequence \cite{funakoshi2025emotion}. 

\item TelME is a conversational emotion recognition architecture that utilizes cross modal distillation to transfer information from a textual teacher to non verbal student, and combines multimodal features through a shifting fusion approach to capture dialogue context \cite{yun-etal-2024-telme}.

\item EmoTrans is a transition-based ERC model and uses recent utterances to capture emotional transitions \cite{jian2024emotrans}.

\item MMGCN is a multimodal graph convolutional network that uses speaker information to model inter-/intra-speaker dependencies across utterances \cite{hu2021mmgcn}.

\end{itemize}

\subsection{Experiment Setup}

\begin{table}[t]
\centering
\renewcommand{\arraystretch}{1.3}
\setlength{\tabcolsep}{9pt}
\caption{Distribution of emotion labels in the processed BOBSL corpus used for weakly supervised post-training.}
\label{tab:bobsl_pretrain}
\begin{tabular}{l c c c}
\toprule
\textbf{Emotion Label} & \textbf{Training Set} & \textbf{Validation Set} & \textbf{Total} \\
\hline
Neutral & 14,962 & 4,992 & 19,954 \\
Joy & 14,929 & 4,990 & 19,919 \\
Sad & 7,429 & 2,545 & 9,974 \\
Angry & 6,211 & 2,386 & 8,597 \\
\hline
Total & 43,531 & 14,913 & 58,444 \\
\bottomrule
\end{tabular}
\end{table}

\begin{table}[t]
\centering
\renewcommand{\arraystretch}{1.3}
\setlength{\tabcolsep}{9pt}
\caption{Distribution of emotion labels in the processed MELD dataset.}
\label{tab:meld_processed}
\begin{tabular}{l c c c}
\toprule
\textbf{Emotion Label} & \textbf{Training Set} & \textbf{Validation Set} & \textbf{Total} \\
\hline
Angry   & 1,109 & 153 & 1,262 \\
Neutral & 4,710 & 470 & 5,180 \\
Joy     & 1,743 & 163 & 1,906 \\
Sad     & 683   & 111 & 794 \\
\hline
Total   & 8,245 & 897 & 9,142 \\
\bottomrule
\end{tabular}
\end{table}

\begin{table*}[t]
\centering
\renewcommand{\arraystretch}{1.2}
\caption{Evaluation results on the eJSL Dialog dataset. The evaluation utilizes the weighted F1 score for overall performance and reports individual F1 scores for the four emotion categories.}
\label{tab:results}
\begin{tabular}{l c l c c c c c c}
\toprule
\textbf{Model} &\textbf{SL data} & \textbf{Modality} & \textbf{Context}& \textbf{Angry} & \textbf{Neutral} & \textbf{Joy} & \textbf{Sad} & \textbf{Weighted F1}\\
\hline
EmoAffectNet & $\checkmark$ & Visual &$\times$& 15.82 & 41.55 & 30.58 & 17.03 & 29.06 \\
EANwH & $\checkmark$ & Visual &$\times$& 9.72 & 55.35 & 35.33 & 13.24 & 33.31 \\
EmoTrans & $\times$ & Text & $\checkmark$ & 44.05 & 57.02 & 72.56 & 42.94 & 55.40 \\
MMGCN &$\times$& Text + Visual & $\checkmark$ & 53.48 & 58.89 & \textbf{82.76} & 40.82 & 60.17 \\
TelME &$\times$ & Text + Visual & $\checkmark$ & \textbf{68.10} & \textbf{66.31} & 82.71 & \textbf{64.04} & \textbf{70.22} \\
\bottomrule
\end{tabular}
\end{table*}

\wy{As EmoAffectNet and EANwH require sign-language visual representations, we followed the weakly supervised post-training strategy of \cite{funakoshi2025emotion} and utilized the BBC Oxford British Sign Language dataset (BOBSL) \cite{Albanie2021bobsl}.
Although BOBSL represents a different linguistic system from our target JSL, both languages share fundamental physical mechanisms for conveying emotions through facial expressions and hand trajectories.
We applied the emotion recognition model \textit{Fine-tuned DistilRoBERTa-base for Emotion Classification}\footnote{\href{https://huggingface.co/michelleli99/emotion_text_classifier}{michelleli99/emotion\_text\_classifier}} to the automatically aligned English subtitles to obtain weak emotion labels, which were used for weakly supervised post-training of the visual baselines.}
We selected 893 videos from the total collection to balance data quality and computational efficiency. 
To align with the emotion categories of the eJSL Dialog dataset, we filtered the samples to retain four emotion labels: Angry, Neutral, Joy, and Sad. 
We resampled the remaining data to achieve a balanced label distribution. 
\wy{\autoref{tab:bobsl_pretrain} details the quantitative distribution of the processed BOBSL corpus used for weakly supervised post-training.}

\wy{For the visual baselines, we applied a consistent preprocessing pipeline to the eJSL Dialog videos.
We uniformly sampled each video clip at 2 frames per second to extract the frame sequence. 
This sampling rate balances temporal resolution and computational efficiency, ensuring that key visual expressions are captured without processing redundant frames.
For facial feature extraction, we used the RetinaFace detector \cite{deng2020retinaface} to identify and crop the face region in each extracted frame. 
The cropped facial images were resized to a spatial resolution of 224 by 224 pixels and converted to RGB format. 
For the hand skeletal feature extraction required by the EANwH, we employed the Wholebody landmarker from the rtmlib library \cite{jiang2023}. 
We extracted 21 spatial keypoints for both the left and right hands. 
The hand keypoint coordinates were mathematically normalized relative to the wrist position and concatenated into an 84-dimensional feature vector for each frame.}

\wy{For models using the text modality, the ground-truth Japanese transcripts of eJSL Dialog were translated into English using Google Translate and directly provided as text input.
Therefore, the text modality in our evaluation should be regarded as oracle information rather than output from an automatic sign-language recognition system.}

\wy{For EmoTrans, TelME, and MMGCN, we retrained the models on the MELD dataset \cite{poria2019meld} using only the four emotion categories shared with eJSL Dialog: Angry, Neutral, Joy, and Sad.
The distribution of the four retained emotion categories in the processed MELD training and validation sets is summarized in \autoref{tab:meld_processed}.
For TelME and MMGCN, since audio is unavailable in eJSL Dialog, the audio features were set to zero during both MELD training and eJSL evaluation.
The resulting models were then evaluated on eJSL Dialog.}
\wy{For the modality settings used in our evaluation, ``Text only'' retains only the oracle textual features and sets the visual features to zero, while ``Visual only'' retains only the visual features and sets the textual features to zero. ``Text + Visual'' provides both textual and visual features. Audio features are set to zero in all settings.}
\wy{To account for the unequal distribution of emotion categories, we use the weighted F1 score as the primary overall metric. 
Furthermore, we report the individual F1 scores for the four emotion categories to analyze specific recognition capabilities.}

\subsection{Evaluation Results}
We present the evaluation results of the baseline models on the eJSL Dialog dataset in \autoref{tab:results}. 
The visual baseline EmoAffectNet achieves a weighted F1 score of 29.06. 
Building upon this, the extended visual model EANwH yields a higher weighted F1 score of 33.31. 
This performance improvement occurs because EANwH incorporates hand skeletal features and utilizes an LSTM network to capture temporal information across the frame sequence. 
\wy{This suggests that sign-language visual features contain affective information that can be recognized without textual input.}
However, recognizing the Sad category remains challenging using only visual cues, with EmoAffectNet recording an F1 score of 17.03 and EANwH recording 13.24.
This suggests that visual-only models are limited by the ambiguity between affective and linguistic visual cues in sign language.

\wy{The text-only model EmoTrans achieves a weighted F1 score of 55.40.
Since the ground-truth transcripts are directly provided as oracle text input, the EmoTrans result represents emotion recognition with access to perfect textual information, rather than end-to-end recognition from sign-language video.
The same oracle-text setting is also used for the multimodal models that take textual input.
Moreover, the dialogue lines from STUDIES dataset used to construct eJSL Dialog are strongly correlated with their assigned emotion labels, which makes textual information particularly informative for this benchmark.
}

\wy{The multimodal conversational models achieve the strongest overall performance under the current evaluation setting.
TelME obtains the highest weighted F1 score of 70.22, while MMGCN achieves 60.17, both outperforming the text-only EmoTrans baseline and the visual-only baselines.
For both models, incorporating visual information further improves performance compared with their corresponding text-only settings.
These results suggest the potential of transferring existing speech-based ERC frameworks to sign-language dialogue.
However, their performance on eJSL Dialog remains lower than that achieved on the corresponding speech-based four-class ERC task \cite{busso2008iemocap,li2023joyful}, indicating room for further improvement in cross-domain adaptation.}

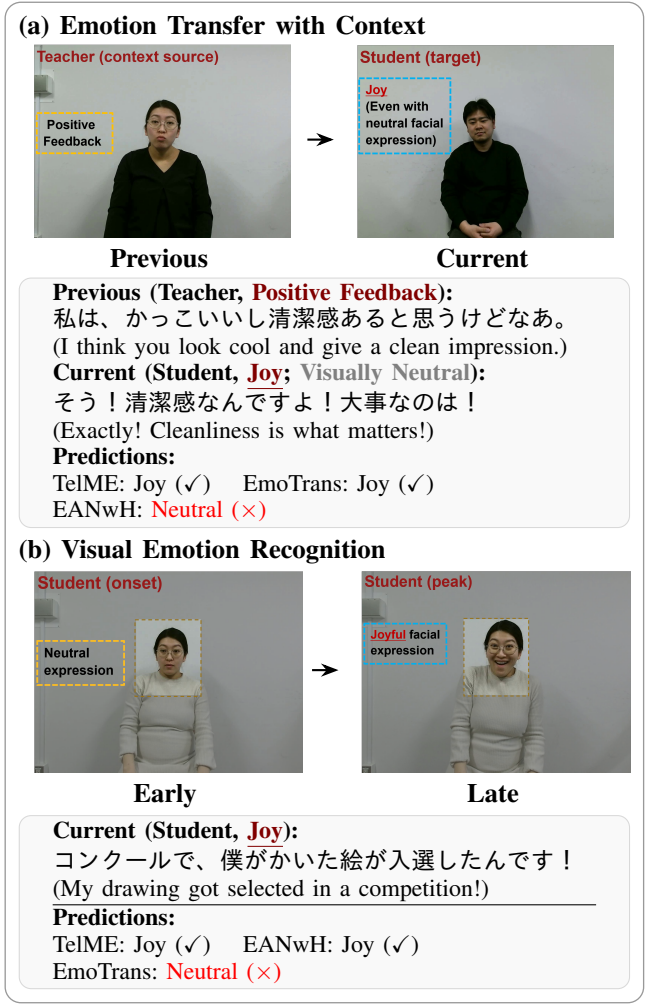
\begin{figure}[t]
\centering

\begin{tcolorbox}[
    colback=white,
    colframe=black!40,
    boxrule=0.5pt,
    arc=2mm,
    width=0.98\linewidth,
    left=2pt,right=2pt,top=2pt,bottom=2pt
]

\textbf{(a) Emotion Transfer with Context}

\vspace{-0.2em}

\begin{tikzpicture}[node distance=0.05cm]

\node (img1) {\includegraphics[width=0.43\linewidth]{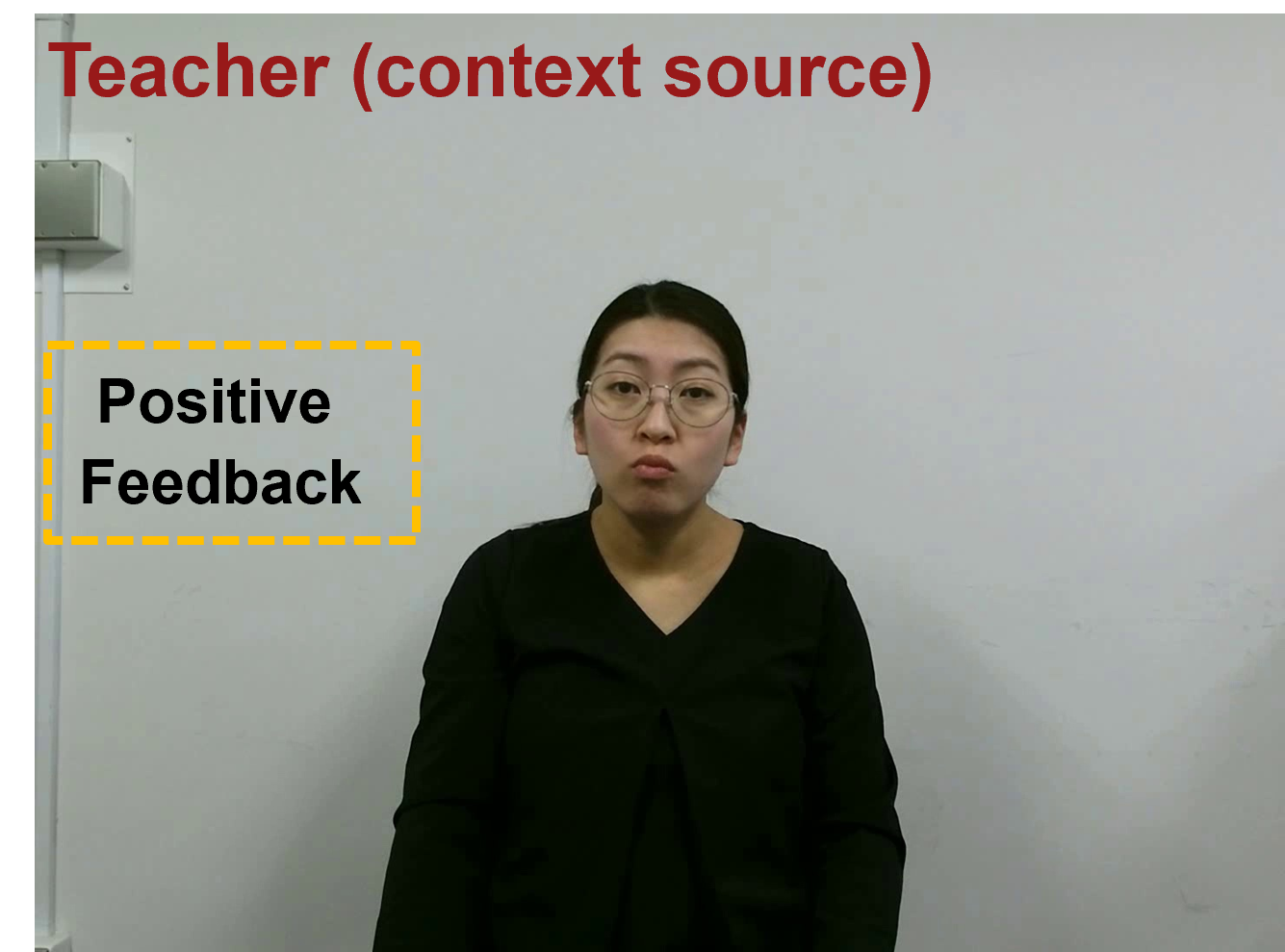}};
\node (img2) [right=of img1, xshift=0.5cm] {\includegraphics[width=0.43\linewidth]{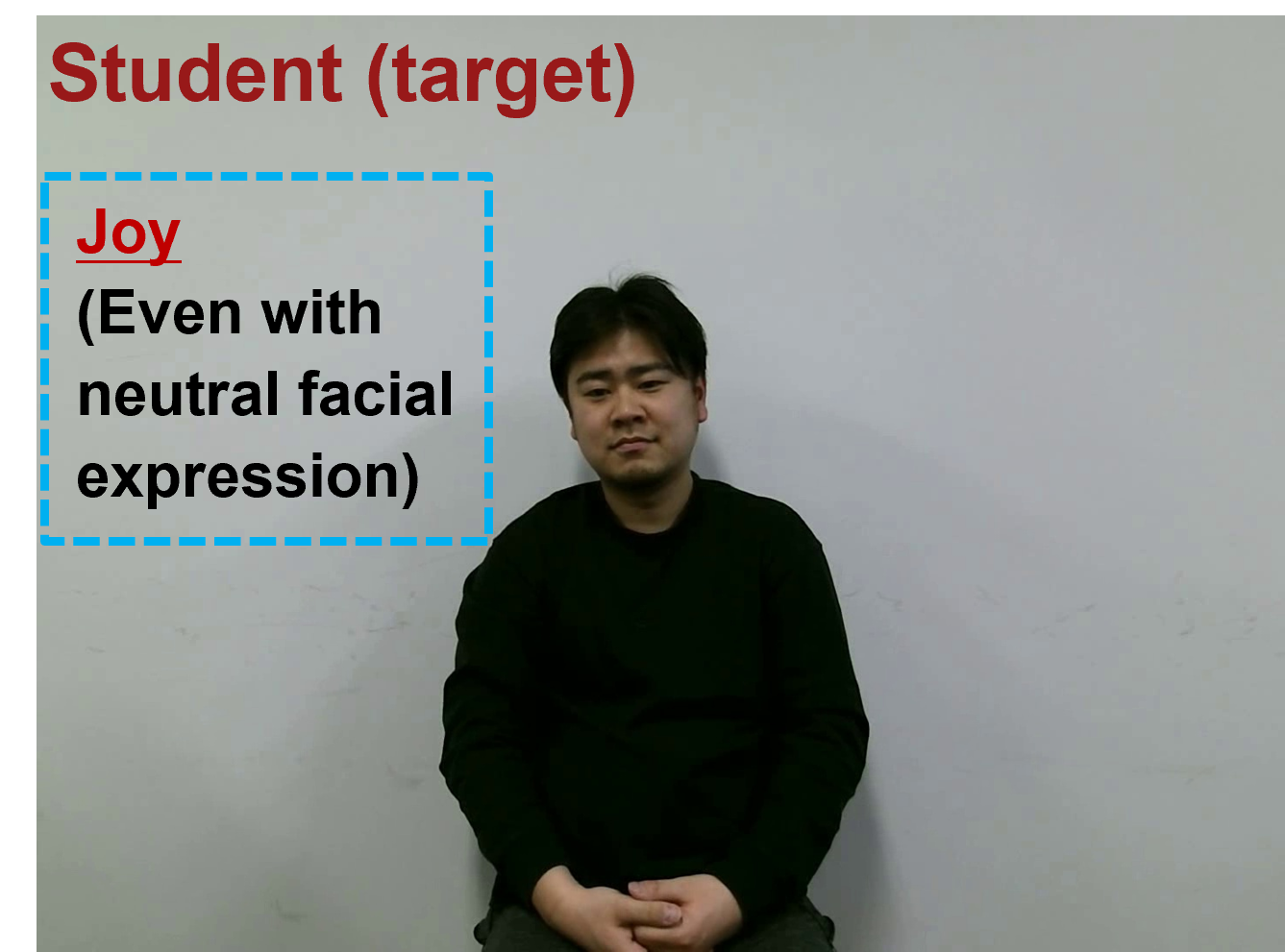}};
\draw[-{Stealth}, thick] 
    ($(img1.east)+(0.1,0)$) -- ($(img2.west)+(-0.1,0)$);
\node[below=-0.1cm of img1] {\textbf{Previous}};
\node[below=-0.1cm of img2] {\textbf{Current}};

\end{tikzpicture}

\vspace{-0.9em}

\begin{tcolorbox}[
colback=gray!5,
colframe=gray!30,
boxrule=0.4pt,
arc=1.5mm,
top=1pt,bottom=1pt,
boxsep=1pt,
before upper={\setlength{\parskip}{0pt}}
]
\small

\textbf{Previous (Teacher, {\color{red!50!black}Positive Feedback}):}  

\ja{私は、かっこいいし清潔感あると思うけどなあ。} \\
(I think you look cool and give a clean impression.)

\textbf{Current (Student, {\color{red!50!black}\underline{Joy}}; {\color{gray}Visually Neutral}):}  

\ja{そう！清潔感なんですよ！大事なのは！} \\
(Exactly! Cleanliness is what matters!)

\textbf{Predictions:} \\
TelME: Joy (\checkmark) \quad
EmoTrans: Joy (\checkmark) \\
EANwH: \textcolor{red}{Neutral ($\times$)}

\end{tcolorbox}

\vspace{-0.3em}

\textbf{(b) Visual Emotion Recognition}

\vspace{-0.2em}

\begin{tikzpicture}[node distance=0.05cm]

\node (img1) {\includegraphics[width=0.45\linewidth]{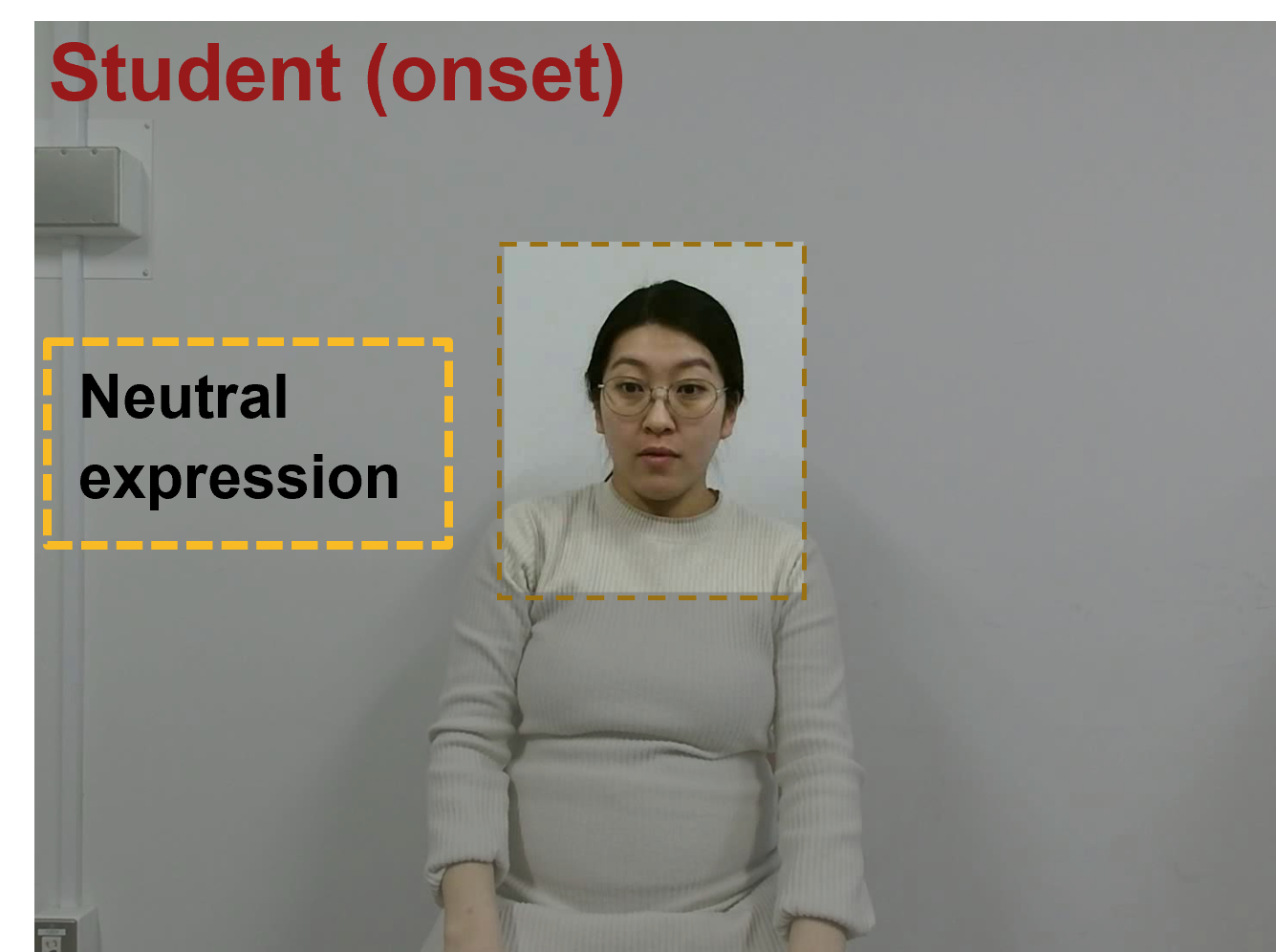}};
\node (img2) [right=of img1, xshift=0.4cm] {\includegraphics[width=0.45\linewidth]{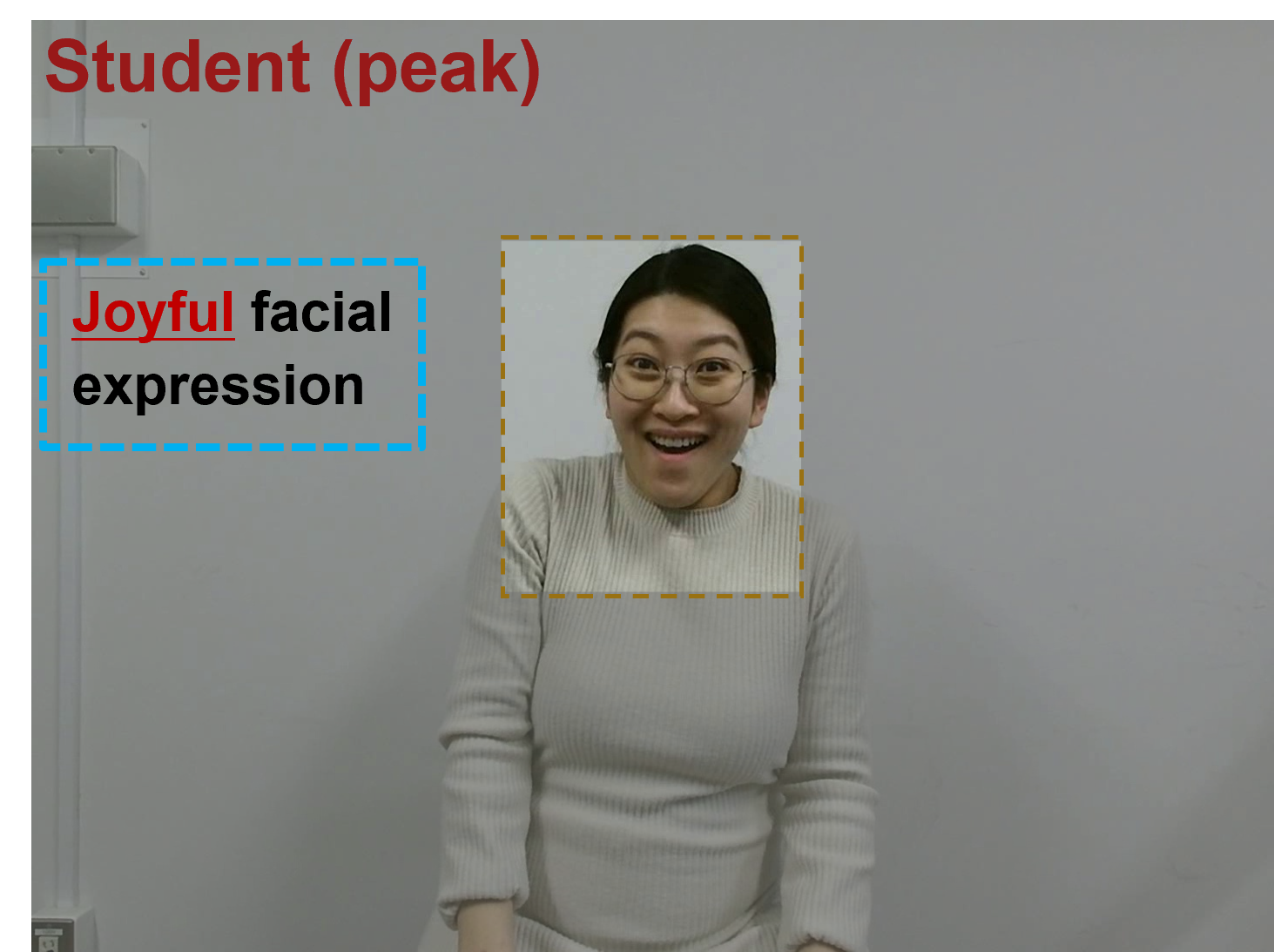}};

\draw[-{Stealth}, thick] 
    ($(img1.east)+(0.,0)$) -- ($(img2.west)+(-0.1,0)$);

\node[below=-0.1cm of img1] {\textbf{Early}};
\node[below=-0.1cm of img2] {\textbf{Late}};

\end{tikzpicture}

\vspace{-1em}

\begin{tcolorbox}[
colback=gray!5,
colframe=gray!30,
boxrule=0.4pt,
arc=1.5mm,
top=1pt,bottom=1pt,
boxsep=1pt,
before upper={\setlength{\parskip}{0pt}}
]
\small

\textbf{Current (Student, {\color{red!50!black}\underline{Joy}}):}   

\ja{コンクールで、僕がかいた絵が入選したんです！} \\
(My drawing got selected in a competition!)

\vspace{0.15em}
\hrule
\vspace{0.15em}

\textbf{Predictions:} \\
TelME: Joy (\checkmark) \quad
EANwH: Joy (\checkmark) \\
EmoTrans: \textcolor{red}{Neutral ($\times$)}

\end{tcolorbox}

\end{tcolorbox}

\caption{
Case studies illustrating complementary roles of conversational context and visual cues. 
(a) Context-aware models correctly capture emotion transfer, while the non-conversational model fails. 
(b) Visual cues enable correct emotion recognition, while the text-only model fails.
}
\label{fig:case_study}
\end{figure}

\subsection{Modality Ablation}

\begin{table*}[t]
\centering
\renewcommand{\arraystretch}{1.2}
\setlength{\tabcolsep}{16pt}
\caption{Modality ablation results of TelME and MMGCN on the eJSL Dialog dataset. Text input corresponds to the oracle transcripts described in the experimental setup.}
\label{tab:modality_ablation}
\begin{tabular}{l l c c c c c}
\toprule
\textbf{Model} & \textbf{Modality} & \textbf{Angry} & \textbf{Neutral} & \textbf{Joy} & \textbf{Sad} & \textbf{Weighted F1} \\
\hline
TelME & Text + Visual & 68.10 & 66.31 & 82.71 & 64.04 & \textbf{70.22} \\
TelME & Text only & 67.99 & 64.98 & 82.09 & 63.56 & 69.47 \\
TelME & Visual only & 0.00 & 53.66 & 0.00 & 0.00 & 19.67 \\
MMGCN & Text + Visual & 53.48 & 58.89 & 82.76 & 40.82 & \textbf{60.17} \\
MMGCN & Text only & 54.96 & 34.45 & 80.39 & 46.61 & 52.03 \\
MMGCN & Visual only & 47.41 & 50.73 & 48.36 & 37.22 & 46.94 \\
\bottomrule
\end{tabular}
\end{table*}

\wy{To further analyze the contributions of textual and visual information in multimodal ERC models, we conduct modality ablation experiments on TelME and MMGCN under text-only, visual-only, and text-visual settings.
As shown in \autoref{tab:modality_ablation}, different ERC architectures exhibit clearly different modality dependence patterns in sign-language dialogue.
TelME is driven primarily by the textual modality, with only a small improvement after incorporating visual information.
In contrast, MMGCN shows relatively limited performance with either text or video alone, but achieves a clear improvement when the two modalities are combined.
These results indicate that the ability of existing ERC models to exploit the complementarity between textual and visual information is strongly architecture-dependent.
Therefore, future research should focus on effectively extracting, modeling, and integrating textual and visual affective information for emotion recognition in sign language conversation.}

\wy{Compared with the context-free visual baselines, context-based methods like the visual-only MMGCN achieve better performance, suggesting that conversational context may provide useful complementary information for visual emotion recognition in sign-language dialogue.
This observation should be interpreted with caution, as differences in preprocessing and visual encoders may also contribute to the performance gap.}

\subsection{Case Study}

To analyze the contribution of conversational and visual information in sign language ERC, we present two representative cases. 
Each case uses a short sign clip, visualized by representative frames to approximate temporal dynamics.

\subsubsection{Case 1: Importance of Conversational Context}
As shown in \autoref{fig:case_study}(a), the student expresses Joy after the teacher's positive feedback. 
Although the facial expressions and manual signs of the student appear restrained, the preceding positive feedback from the teacher, combined with the affirmative and excited textual semantics of the student utterance, jointly enable the text based model EmoTrans and the multimodal model TelME to make the correct Joy prediction. 
In contrast, the visual model EANwH, which uses only current visual frames and does not model conversational context, fails to capture this transition and incorrectly outputs a Neutral prediction. 
This demonstrates that contextual reasoning and textual semantics are critical for modeling emotion dynamics in dialogue based sign language scenarios.

\subsubsection{Case 2: Importance of Visual Cues}

As shown in \autoref{fig:case_study}(b), the emotion can convey through the signer's facial expressions and visual cues rather than textual content alone. 
Although the text describes a positive event, this factual statement is misinterpreted by the language model as a neutral report because it lacks explicit emotional modifiers. 
The multimodal model TelME and the visual model EANwH successfully capture these dynamic visual signals and correctly classify the emotion as Joy. 
This highlights the importance of visual information in sign language emotion recognition, where facial expressions play a crucial role in confirming the true affective state of the speaker.

Overall, these cases reveal three complementary factors in sign language emotion understanding: conversational context, textual semantics, and visual cues. 
Conversational context provides the necessary history for tracking emotional shifts across turns. 
Furthermore, while textual semantics help capture implicit emotion transitions when visual expressions are restrained, visually grounded signals remain essential for resolving ambiguity in factual textual expressions.
Integrating these three elements into a unified multimodal architecture offers a clear direction for further improvement.

\section{IMPACT, LIMITATIONS, AND FUTURE WORK}

\paragraph{Impact}
In real sign language conversation, emotions evolve across conversational turns. 
By providing a benchmark for bidirectional interactions, this research supports the future development of empathetic dialogue systems that can track and respond to the emotional history of deaf users. 
This capability is a requirement for creating virtual assistants that can engage in natural and context aware human computer interaction for deaf people.

\paragraph{Limitations}
We identify several limitations in the current study. 
Regarding the data scope, the eJSL Dialog dataset features only two actors and was recorded in a controlled laboratory environment with a pure white background.
Although eJSL Dialog is used as a benchmark rather than a training set, the baseline results may still be affected by signer-specific appearance, signing style, and expressive habits.
Consequently, the current results should be interpreted as an initial evaluation under limited signer diversity, and models positively evaluated on this dataset might still struggle with generalization when applied to in-the-wild environments that contain diverse backgrounds, lighting conditions, and signer demographics.
The emotion categories are also constrained by the original STUDIES corpus. 
We adopted its four labels to preserve script-label consistency and class balance, but this limits coverage of broader affective states in sign-language communication.
Moreover, eJSL Dialog does not provide separate annotations for emotional facial cues and linguistic non-manual markers, and the evaluated baselines do not explicitly disentangle these two types of visual signals. 
Therefore, the current benchmark cannot determine whether models distinguish linguistic markers from emotional states.
\wy{In addition, the text-based and multimodal models use ground-truth transcripts as oracle text input.
At present, a reliable and readily implementable Japanese Sign Language recognition model is not available for our experimental setting, so we directly provide the ground-truth transcripts to isolate the evaluation of conversational emotion recognition from sign-language recognition errors.
Therefore, the current evaluation does not account for errors introduced by automatic sign-language recognition or translation, and the reported text-related performance should not be interpreted as end-to-end sign-language emotion recognition performance.}

\paragraph{Future Work}
\wy{Future work will address these limitations by expanding signer diversity, collecting longer dialogue sessions, and incorporating spontaneous sign-language conversations.
We will also construct larger-scale sign-language conversational training datasets to support the development of models that can more effectively extract, model, and integrate textual and visual affective information across dialogue turns.
In addition, future evaluation will replace oracle transcripts with automatically recognized text or glosses to investigate emotion recognition under more realistic end-to-end settings.}

\section{CONCLUSIONS}
We introduced the ERC task for sign language and proposed the eJSL Dialog dataset. 
By providing multiple turn dialogue history, this dataset addresses the limitations of isolated sign language emotion recognition.
\wy{Our benchmark results suggest the potential of transferring existing speech-based ERC frameworks to sign-language dialogue, while their remaining performance gap on eJSL Dialog indicates clear room for further improvement in cross-domain adaptation.
The modality ablation further reveals different modality-dependence patterns across ERC architectures, highlighting the importance of effectively extracting, modeling, and integrating textual and visual affective information for sign-language dialogue.}




\section*{ACKNOWLEDGMENT}
The construction of eJSL datasets was supported by Tateisi Science and Technology Foundation (Representative: Rei Kawakami, 2025).
This work was supported by Japan Science and Technology Agency (JST) as part of Adopting Sustainable Partnerships for Innovative Research Ecosystem (ASPIRE), Grant Number JPMJAP25B3.


\bibliographystyle{IEEEtran}
\bibliography{IEEEabrv,refs}

\end{document}